\documentclass[letterpaper, 10 pt, conference]{ieeeconf} 
\IEEEoverridecommandlockouts  
\usepackage{cite}
\usepackage{amsmath,amssymb,amsfonts}
\usepackage{algorithmic}
\usepackage{graphicx}
\usepackage{caption}
\usepackage{subcaption}

\usepackage{float}
\usepackage{hyperref}
\usepackage{mathtools} 
\usepackage{textcomp}
\usepackage{xcolor}
\usepackage{multirow}
\usepackage{makecell} 

\def\BibTeX{{\rm B\kern-.05em{\sc i\kern-.025em b}\kern-.08em
    T\kern-.1667em\lower.7ex\hbox{E}\kern-.125emX}}

\begin{document}

\title{Input Shaping for Point-to-Point Motion with a Continuum Robot Arm}

\author{Rodolfo Hdz. Ibarra, Karan Baker, Parsa Molaei, Adrian Stein, and Hunter B. Gilbert
\thanks{R. H. Ibarra, K. Baker, P. Molaei, A. Stein, and H. B. Gilbert are with the Department of Mechanical and Industrial Engineering, Louisiana State University, LA 70803, USA. {\tt\small(email: \{rherna10,kbake54,pmolae1,astein,hbgilbert\} @lsu.edu)}.}
\thanks{This material is based upon work supported in part by the National Science Foundation under Grant Number 2133019.}}

\newcommand{\hbg}[1]{\textcolor{red}{[HBG: #1]}}
\newcommand{\pmo}[1]{\textcolor{blue}{[PMO: #1]}}

\maketitle

\begin{abstract}
A cable-driven continuum robot arm is an underactuated mechanism and may suffer residual vibration at the end of a rest-to-rest maneuver. In this work, a time-delay filter is applied as an input shaper to the system to eliminate the excitation of vibratory modes. A non-robust and a robust time-delay filter are designed based on a linear system model and demonstrate improved response compared to a velocity-driven pulse input. Experimental results using the continuum robot validate the application of the input shaper, with reduced overshoot and settling time exemplifying the reduction in oscillation at the end of the maneuver. It is also shown that utilizing the robust shaper further improves the response of the arm in comparison to applying the non-robust shaper. These results are significant towards the precise and robust implementation of continuum robots in  applications involving arbitrary end-effector trajectories. 
\end{abstract}

\begin{keywords}
Input Shaping, Precision Motion Control, Continuum Robot, Vibration Control
\end{keywords}
%
%
%
\section{INTRODUCTION}
\label{sec:introduction}
Slender-bodied, mechanically flexible, snake-like robots have a wide variety of uses in inspection, medicine, and other practical applications where access to confined spaces and complex environments is facilitated by a body that is passively compliant~\cite{burgner-kahrs_continuum_2015,rus_design_2015}, or in general where a passively compliant and lightweight body is desirable~\cite{davies_subsea_1998}. These robots, in contrast to traditional robotic mechanisms composed of kinematic linkages of nearly-rigid bodies, are usually underactuated (meaning that they do not have enough actuators to fully control their own body shape), and they are typically compliant enough that expected external forces on the body of the robot cause deformation that is significant~\cite{gilbert_mathematical_2021}. 

\begin{figure}
\centering
\includegraphics[width=3.4in]{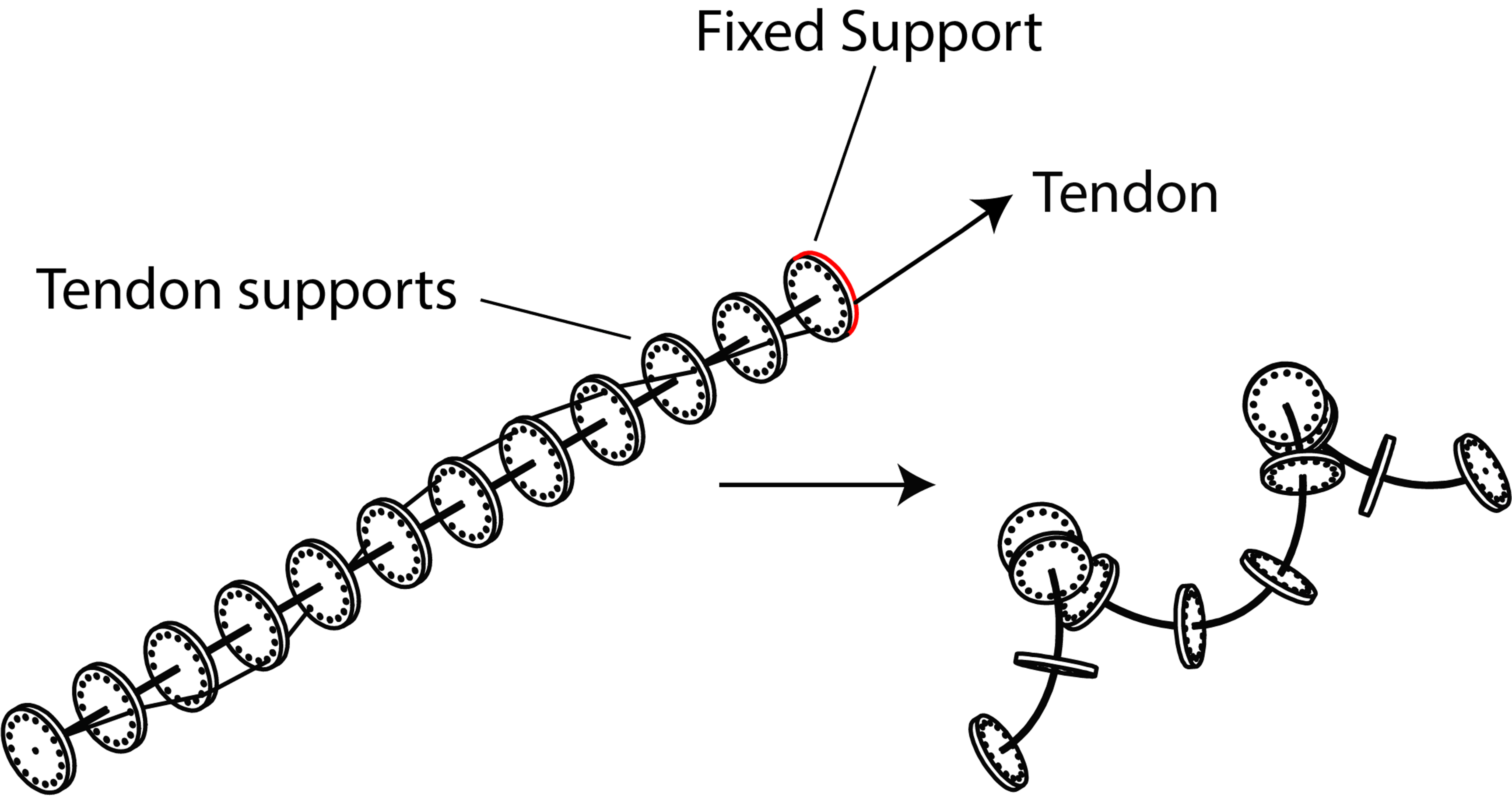}
\caption{Diagram illustrating how cable-driven continuum robots use cable tension to cause shape changes in elastic structures.}
\label{fig:1_Robot_Drawing}
\end{figure}

A variety of designs for continuum robots have been proposed. For example, some use pneumatic artificial muscles~\cite{trivedi_optimal_2008}, and some use electromechanical actuators such as motor-driven parallel elastic push-pull rods~\cite{simaan_design_2009}, concentric pre-curved tubes~\cite{gilbert_concentric_2016}, and flexible wire ropes or cables--the latter also referred to as ``tendons.'' A typical structure for cable-driven continuum robot arms is shown in Fig.~\ref{fig:1_Robot_Drawing}, where an actuating cable is routed through rigid supports. As the cable is pulled, less of its length lies along the path defined by the eyelets in the cable support discs, and the elastic backbone of the robot bends under the loads applied by the cable.

The cable has a dual effect on the elastic backbone. First, it provides a generalized force that causes the robot to bend and/or twist. The resulting shape depends on the routing of the cable through the supports and the elastic properties of the backbone. Second, it modifies the structural rigidity along the shape mode(s) that the cable couples to~\cite{molaei_cable_2022, molaei_independent_2023, oliver-butler_continuum_2019}. Pre-tension in antagonistic cables can actively soften the structure~\cite{peyron_stability_2024}. The elastic structure means that high-bandwidth input signals in the cable displacement cause vibration, and the structural rigidity modification of the cable means that the vibrational characteristics are also altered by the cable depending on its route and tension. 
The modeling of continuum robots advanced from an understanding captured by static models to linear dynamic models and eventually nonlinear dynamic models based on rods \cite{webster_design_2010,rucker_statics_2011}. Control is often implemented at the kinematic level, with resolved-rate methods based on the system's velocity coefficients (manipulator Jacobian matrix) used to calculate actuator input rates that will accomplish desired task-space motions \cite{yip_model-less_2014}. From the standpoint of the state variables of the robot body, the control of cable-driven continuum robots is often open-loop, with feedback provided by a model or human operator, but can also be closed-loop on the robot shape using suitable shape sensors \cite{bicchi_model-based_2025}. Dynamic feedback control designs have often been model-based and include PD-plus-feedforward type control \cite{gravagne_uniform_2002}, feedback linearization \cite{deutschmann_position_2017,rucker_task-space_2022}, and learning-based controllers \cite{braganza_neural_2007,maghooli_learning-based_2025}, among others. 

In open-loop rate-controlled applications such as teleoperated systems for medicine or industrial inspection and maintenance, arbitrary input commands can excite the flexible-body vibration modes of the robot, leading to significant degradation in position or velocity tracking during both operator-commanded and pre-programmed motions. To address this challenge, we apply an input shaping strategy in the form of a time-delay filter (TDF)~\cite{singh_optimal_2009,stein_minimum_2023}. A TDF operates by placing zeros on the system poles associated with vibratory modes, thereby attenuating their effect. Because the filter simply splits and delays the original input signal, it can be seamlessly integrated into open-loop control architectures. Moreover, robustness can be introduced by placing an additional pair of zeros at the same poles' location, which ensures that even under slight frequency uncertainties, vibration reduction remains effective. This enhanced design will be referred to as the robust TDF throughout this work.

The work is organized as follows. Section~\ref{sec:methodology} details experimental setup of the continuum robot, the system identification, and the input shaper design. Section~\ref{sec:results} presents results of the experiments where different input shaper design are compared with each other for various settings. Section~\ref{sec:conclusions} concludes and discusses future research directions.
%
%
%
\section{METHODOLOGY}
\label{sec:methodology}
This section describes the experimental setup, system identification, and design of input shapers.
%
%
%
\subsection{Experimental Setup}
\begin{figure}
    \centering
    \includegraphics[width=0.45\textwidth]{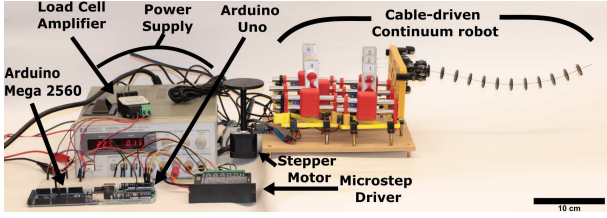}
    \caption{Continuum robot setup including Arduino Mega (data acquisition), Arduino Uno (motor control), TB$6600$ driver, stepper motor (MOTOU $17$HS$4401$S), and LCA-RTC amplifier.}
    \label{fig:2_Continuum_robot_setup}
\end{figure}

A cable-driven continuum robot was developed using load cell sensors (LSP-5, range: $0$–$5$N, Transducer Techniques, Temecula, CA, USA) connected through a load cell amplifier (LCA-RTC, Transducer Techniques). The amplifier converts the millivolt-bridge signals produced by the sensors into measurable outputs for further processing. The complete experimental setup, including control electronics and power supply, is illustrated in Fig.~\ref{fig:2_Continuum_robot_setup}. 
The LCA-RTC amplifier transmitted its output to an Arduino Mega 2560. A $150 \Omega$ resistor was implemented to convert the amplifier’s current-loop output signal of $0.6$ mA into a measurable voltage. The analog signal was read through one of the Arduino's built-in $10$-bit ADC inputs. Power is provided by an adjustable benchtop power supply ($11.8$-$26$ V, Keysight E3630A). 

A second microcontroller (Arduino Uno) is used to control a stepper motor (MOTOU 17HS4401S, NEMA 17, $1.8$° step angle, $4.3$ kg·cm holding torque), which provides input commands via digital logic output to a stepper motor driver (TB6600, Maker Group Global LLC). The motor setup is $800$ steps/revolution.

The physical components of the robot are mounted to a single ground plate (see Fig~\ref{fig:2_Continuum_robot_setup}), which is always fixed in the inertial lab frame of reference. The stepper motor is mounted to the ground plate and turns a cable spool with a diameter of $6.5$ cm. One end of the cable is affixed to the spool, and the other is affixed to one end of a load cell sensor. The load cell sensor slides on a pair of linear rails. On the other terminal of the load cell sensor, a second cable originates. The second cable is routed through a series of ball-bearing v-groove pulleys and into the proximal end of the continuum arm. The cable is routed through the arm and located relative to an elastic backbone by $12$ cable-support discs that are rigidly affixed to the backbone. The second cable terminates at the final support disc with a non-slipping knot that ties it to the final support disc. The backbone has a uniform circular cross-section with a diameter of $0.82$ mm and is made of A228 music wire. The total length of the robot is $L=0.24~\text{m}$

\begin{figure}
    \centering
    \includegraphics[width=0.35\textwidth]{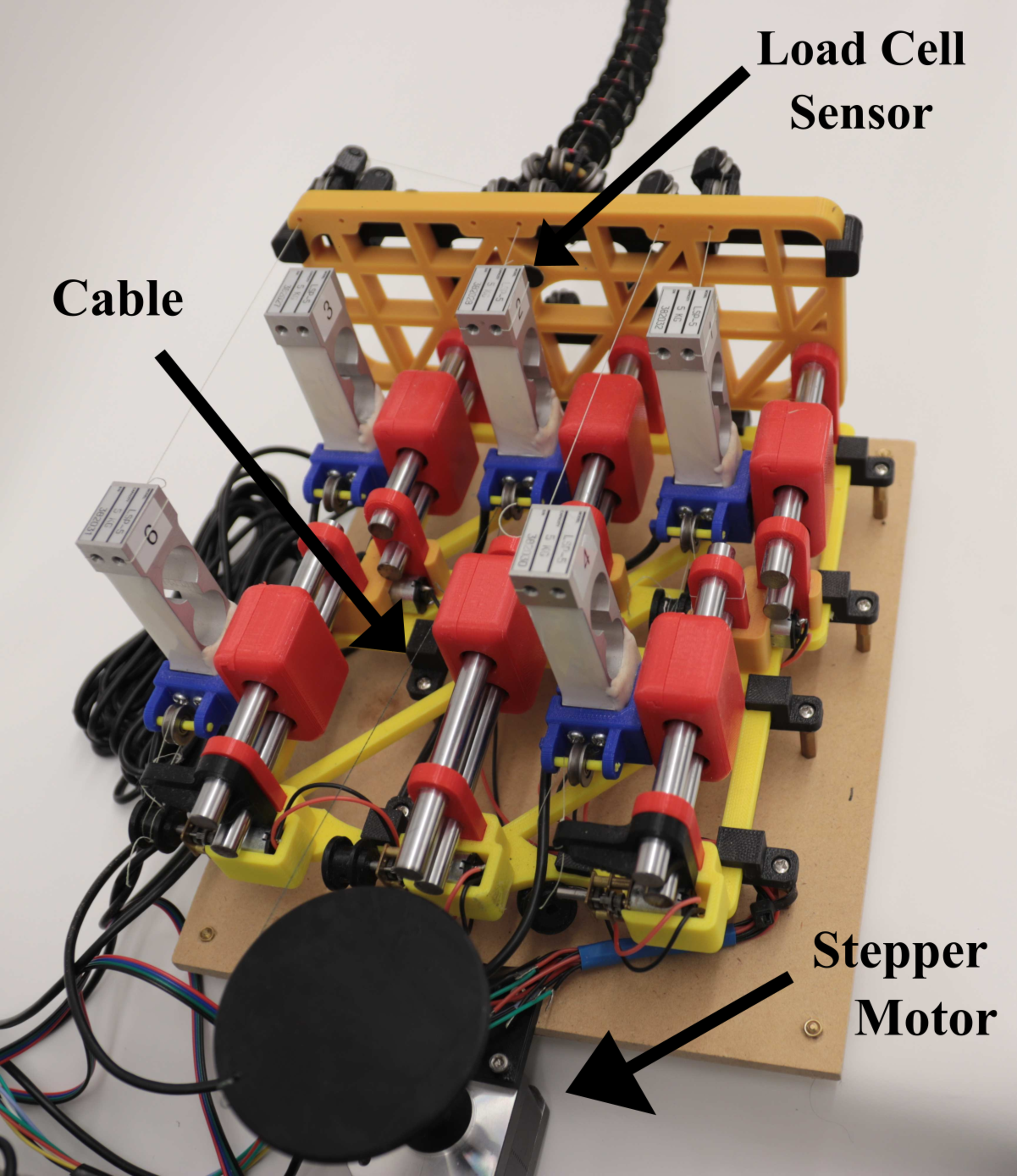}
    \caption{Back view of the continuum robot body, showing the cable routing, load cell sensor, and stepper motor responsible for cable actuation.}
    \label{fig:3_continuum_robot_backbody}
\end{figure}

To better illustrate the mechanical integration, Fig.~\ref{fig:3_continuum_robot_backbody} shows the back view of the continuum robot body. This highlights the arrangement of the load cell sensor, the cable routing, and the stepper motor used for actuation. The apparatus was originally developed to facilitate control of many degrees of freedom; however, in this work only one degree of freedom is actuated.
%
%
%
\subsection{System Identification}
The dynamic model of the continuum arm is assumed to have the cable velocity $v_{in}(t)$ as the causal input. Models for the internal state of the system vary in complexity, but in principle a highly accurate model might involve state variables such as the distributed stretch in the cable and the local curvature and torsion $\kappa(x)$ and $\tau(x)$ of the arm (measured in radians per meter), where $x$ is the arc-length (measured in meters) measured along the geometric centerline of the slender arm \cite{gilbert_mathematical_2021}. The position of the tip of the robot often serves as an output quantity of interest. However, for the purpose of designing the input shaper, the cable tension $\lambda(t)$ is a particularly useful output because vibrations of flexible modes that couple strongly to the actuator, which are also those that can be best controlled by the actuator, are reflected in the tension. The tension is also highly correlated with the system state and the tip position (a fact illustrated later by determining the transfer functions for each and examining the output responses). 

Most state-of-the-art models of continuum arms are geometrically nonlinear models such as the Cosserat rod model. For the purposes of the present study, we consider a linearized plant model which consists of the first $N$ independent modes of vibration and an algebraic equation of constraint created by the cable (assuming that the cable has finite extensional stiffness). The constraint of the cable involves the generalized coordinates, the cable tension, and the imposed cable displacement. Let $\lambda$ be the cable tension and $u$ be the distance the cable is pulled. Then kinematic compatibility requires that
\begin{align} 
    \sum_{n=1}^N K_n q_n + \dfrac{\lambda}{k} - u &= 0.
\end{align}
The parameters $K_n$ are the mode coupling coefficients that describe how the total length of cable that is routed through the robot changes with respect to the modal coordinates $q_n$, which describe the amount of displacement along each of the orthonormal eigenmodes. 
The equations of motion are assumed to be independent from one another with the cable tension forcing each via the coupling coefficient:
\begin{equation}
\begin{aligned}
m_1 \ddot{q}_1 + c_1 \dot{q}_1 + k_1 q_1 &= \lambda K_1, \\
\vdots
\\
m_N \ddot{q}_N + c_N \dot{q}_N + k_N q_N &= \lambda K_N.
\end{aligned}
\end{equation}
For the standard Euler-Bernoulli beam model with uniform properties along the length of the cantilevered beam, the orthonormal mode functions $\phi_n$ are as follows:

\begin{equation}
\begin{split}
\phi_n(x) &= \cosh(\beta_n x/L) - \cos(\beta_n x/L) \\ 
&- \sigma_n \left( \sinh(\beta_n x/L) - \sin(\beta_n x/L) \right)
\end{split}
\end{equation}

\begin{equation}
\sigma_n = \dfrac{\cosh(\beta_n) + \cos(\beta_n)}{\sinh (\beta_n) + \sin(\beta_n)}
\end{equation}
Assuming a uniform parameter distribution for the mass density, the modal masses are identical, $m_n = m/4$ where $m$ is the total mass. The mode characteristic roots $\beta_N$ are ascending in value with mode number. For the cantilevered beam, the first four values of $\beta_n$ are 1.875, 4.694, 7.855, and 10.996. The modal stiffnesses may be expressed in relation to the mass and natural frequency of each mode as $k_n = m_n \omega_n^2$, where
\begin{equation}
    \omega_n = \beta_n^2 \sqrt{\dfrac{EI}{\rho A L^4}}.
\end{equation}
The coupling coefficients depend on the radius of the cable from the neutral axis of bending of the robot and are calculated as 
\begin{equation}
    K_n = r\int_0^L \phi''_n(x) dx = r\phi'_n(L)
\end{equation}
The first four values of $K_nL/r$ are 1.3765, 4.7811, 7.8470, and 10.9913.

The damping parameters $c_1$ through $c_N$ are determined based on a Rayleigh damping model that sets the first two modes to the same effective damping ratio $\zeta$. The relation is of the form
\begin{align}
    c_n &= \eta m_n + \delta k_n, \\
    \delta &= \dfrac{2\zeta}{\omega_1+\omega_2}, \\
    \eta &= \dfrac{2\zeta \omega_1 \omega_2}{\omega_1 + \omega_2}.
\end{align}
The following transfer function is found for the truncated modal model:
\begin{align}
\dfrac{\Lambda(s)}{sU(s)} &= \dfrac{1}{s(\frac{1}{k} + \sum_{n=1}^N{\frac{K_n^2}{D_n}} )}, 
\label{eq:truncated_modal_model_v1}\\
D_n &= m_n s^2 + c_n s + m_n \omega_n^2 \quad \forall n=1,2,...
\label{eq:truncated_modal_model_v2}
\end{align}
The system model order depends on the number of terms retained, but is in general $2N+1$. The transfer function is strictly proper except in the limiting case of $k\to \infty$, which causes the transfer function to be improper. 

The transfer function relating to the displacement of the robot end effector is as follows:
\begin{equation}
\dfrac{Y(s)}{sU(s)} = \dfrac{\sum_{n=1}^N \frac{K_n}{D_n}}{s(\frac{1}{k} + \sum_{n=1}^N{\frac{K_n^2}{D_n}} )}.
\end{equation}
In general, because of the decrease in coupling between the work done by the cable and the energy stored in modes of increasing index, it may be expected that the dominant poles will be not too far from the first mode  \cite{molaei_continuously_2024}. A numerical test using appropriate physical parameter values yields a dominant response mode with a damped natural frequency that is between the first and second modal frequencies. 

\begin{figure}
    \centering
    \includegraphics[width=0.9\linewidth]{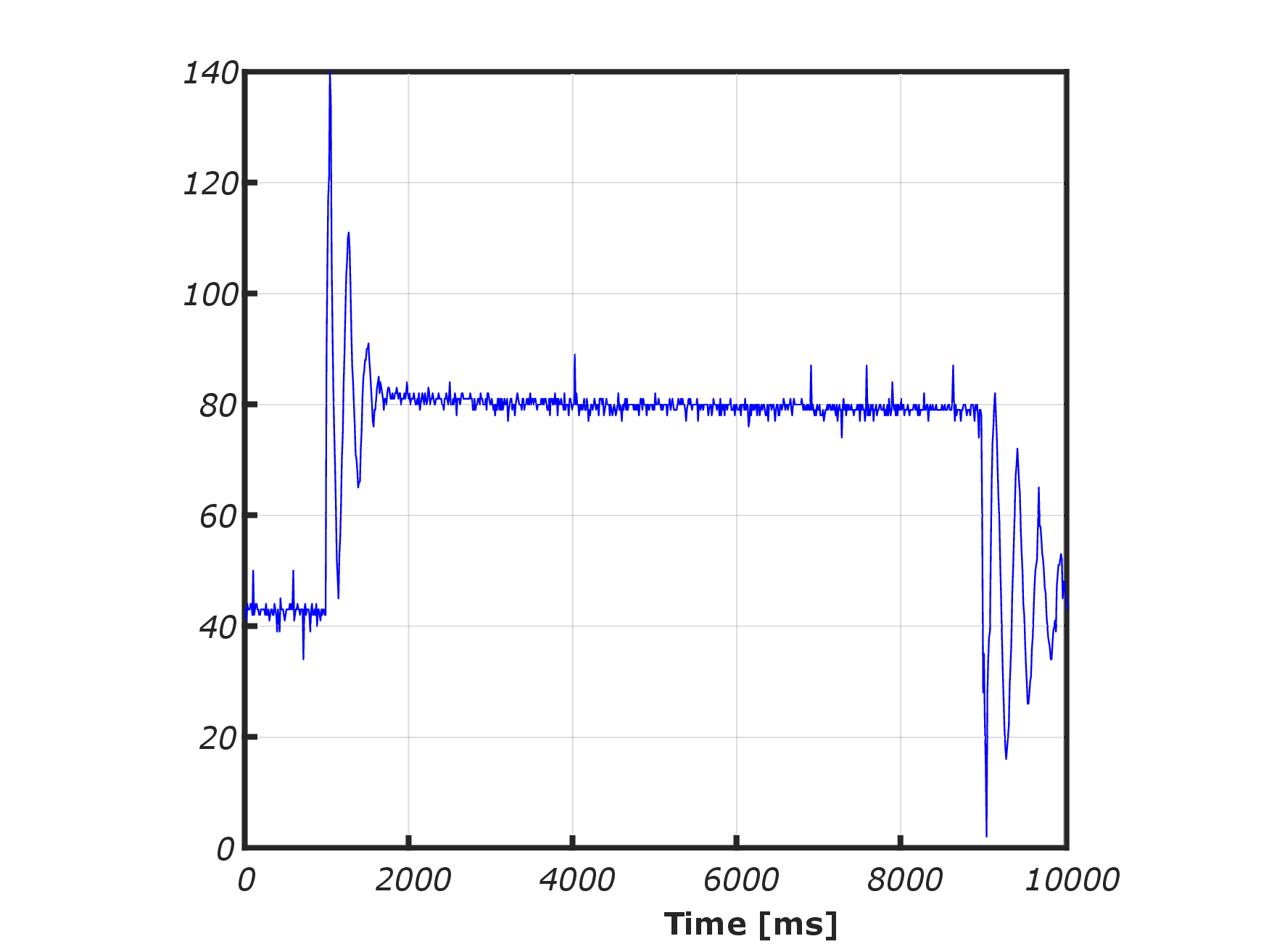}
    \caption{Force response of the continuum robot under pulse input. The signal shows the initial pull of the cable, a steady upward position, and the release. The vertical axis is measured in uncalibrated counts of the analog-to-digital converter.}
    \label{fig:5_Force_Response}
\end{figure}

\begin{figure}
    \centering
    \includegraphics[width=0.9\linewidth]{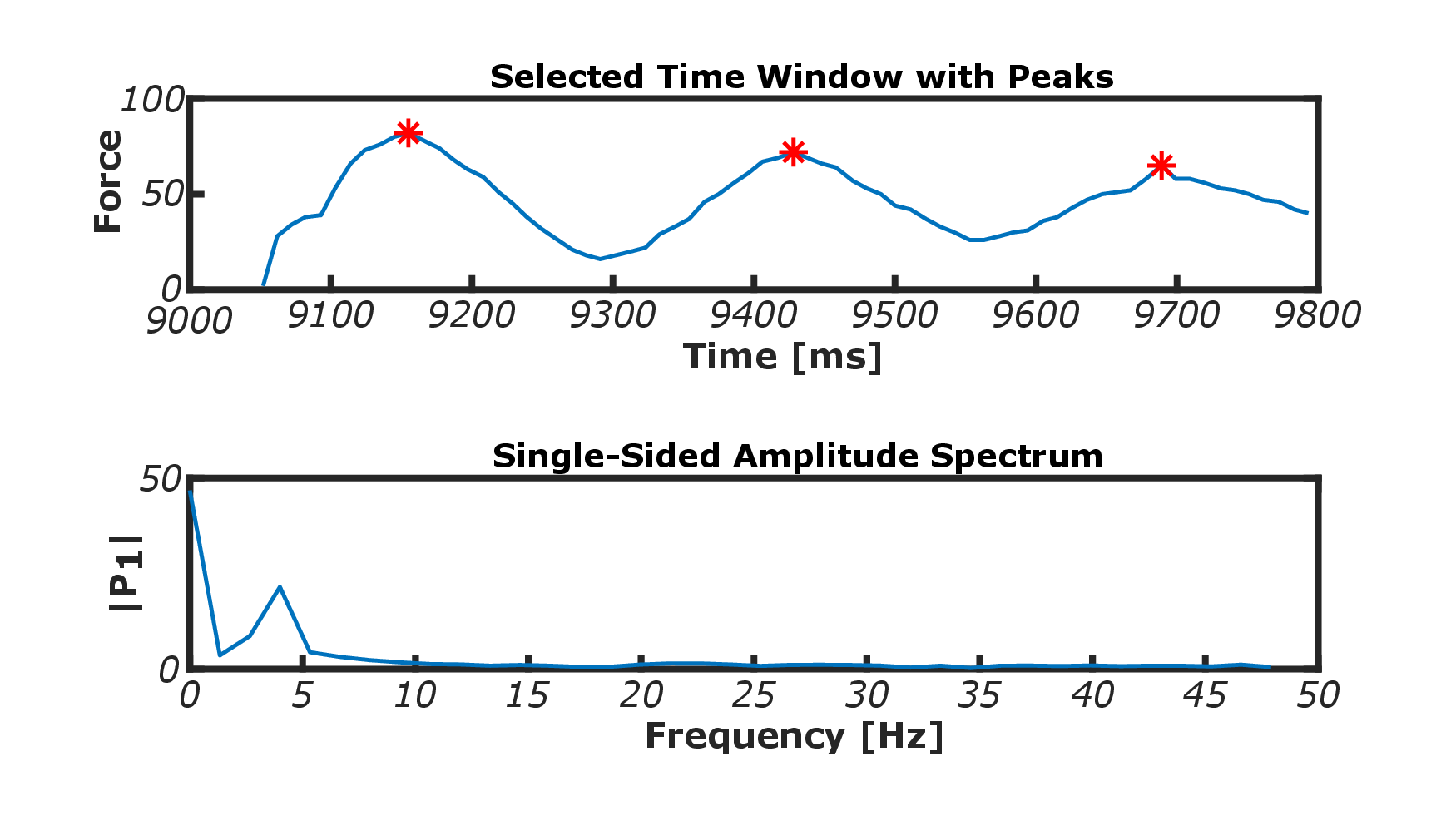}
    \caption{Frequency-domain analysis of the release motion. The upper plot shows the selected time window with detected peaks used to estimate damping. The lower plot presents the single-sided amplitude spectrum obtained from the FFT, highlighting the dominant vibration frequency of the system.}
    \label{fig:6_Log_decrement_and_FFT}
\end{figure}

To identify the system parameters, in particular the dominant poles, a series of velocity pulses are provided to the system. One pulse brings the robot into the up-position and the other pulse back to the down-position. 
Initially the system is at rest in a down-position. The velocity profile can be written as:
\begin{align}
    \dot{u}(t)=\begin{cases}
        v_{in} & 1\leq t < 1+t_p\\
        0 & 1+t_{p} \leq t < 8 \\
        -v_{in} & 8 \leq t < 8+t_p 
    \end{cases},
\end{align}
where $t_p$ the pulse time which results from $t_p=x_f/v_{in}$ with $x_f$ being the final displacement of the pulling cable of the stepper motor. This input generates two distinct transient responses in the system which are used to identify the pole locations. A single response was analyzed as shown in Fig.~\ref{fig:5_Force_Response}. The time between $9$ and $10$ s is used for the system identification. A velocity of $100$ mm/s and a displacement of $x_f=10$ mm were applied.
Fig.~\ref{fig:6_Log_decrement_and_FFT} illustrates the log decrement and fast fourier transform on the last portion of the graph ($9$-$10$ s) presented in Fig.~\ref{fig:5_Force_Response}. From the pulse-release transient, a lightly damped mode with $\zeta=0.044$ and $\omega_d=25.085~\mathrm{rad/s}$ is identified. 
In addition to the identification trial, further cases are selected to systematically evaluate the influence of actuation speed and commanded displacement. The input velocity $v_{in}$ and displacement $x_f$ define the corresponding pulse time $t_p$, which determines the duration of the pull and release scenarios. Each experiment begins with the system at rest in the down position, followed by an upward pull of length $t_p$, a hold phase, and a symmetric return to the initial configuration. The study considers displacements of $10$, $20$, and $30$ mm combined with actuation speeds of $200$ mm/s and $300$ mm/s. Table~\ref{tab:pulse_cases} lists the selected $v_{in}$ and $x_f$ values, along with $t_p$.
\begin{table}
\centering
\caption{Experimental scenarios with input velocity $v_{in}$, displacement $x_f$, and pulse time $t_p$.}
\renewcommand{\arraystretch}{1.4}%
\setlength{\tabcolsep}{6pt}%
\begin{tabular}{|l|c|c|c|}
\hline
\textbf{Profile} & $v_{in}$ (mm/s) & $x_f$ (mm) & $t_p$ (s) \\
\hline
\multirow{4}{*}{Pulse}          & $200$ & $10$ & $0.05$  \\ \cline{2-4}
                                & $200$ & $20$ & $0.10$  \\ \cline{2-4}
                                & $200$ & $30$ & $0.15$  \\ \cline{2-4}
                                & $300$ & $20$ & $0.067$ \\ \hline
\multirow{4}{*}{Non-robust TDF} & $200$ & $10$ & $0.05$  \\ \cline{2-4}
                                & $200$ & $20$ & $0.10$  \\ \cline{2-4}
                                & $200$ & $30$ & $0.15$  \\ \cline{2-4}
                                & $300$ & $20$ & $0.067$ \\ \hline
\multirow{4}{*}{Robust TDF}     & $200$ & $10$ & $0.05$  \\ \cline{2-4}
                                & $200$ & $20$ & $0.10$  \\ \cline{2-4}
                                & $200$ & $30$ & $0.15$  \\ \cline{2-4}
                                & $300$ & $20$ & $0.067$ \\ \hline
\end{tabular}
\label{tab:pulse_cases}
\end{table}
%
%
%
\subsection{Input Shaping}
\label{sec:input_shaping}
\begin{figure}
\centering
	\includegraphics[width=3.2in]{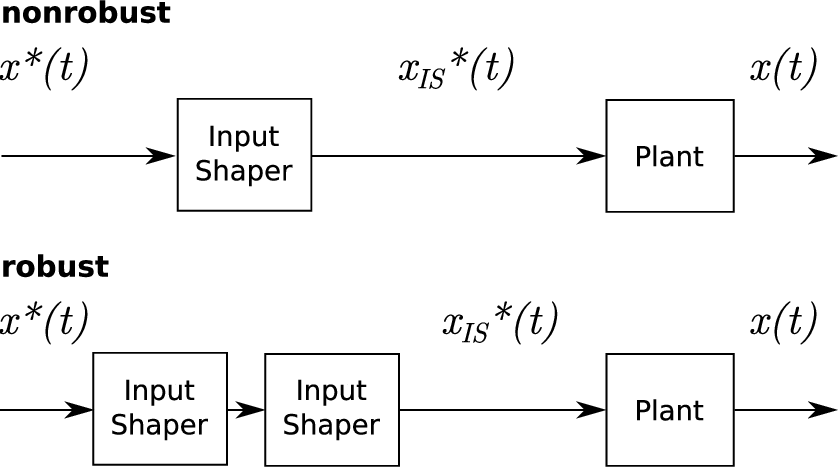}
	\caption{Block diagram of different input shaper designs.}
 \label{fig:4_Block_Diagram_Input_Shaping}
\end{figure}

\begin{figure}
    \centering
    \includegraphics[width=3.2in]{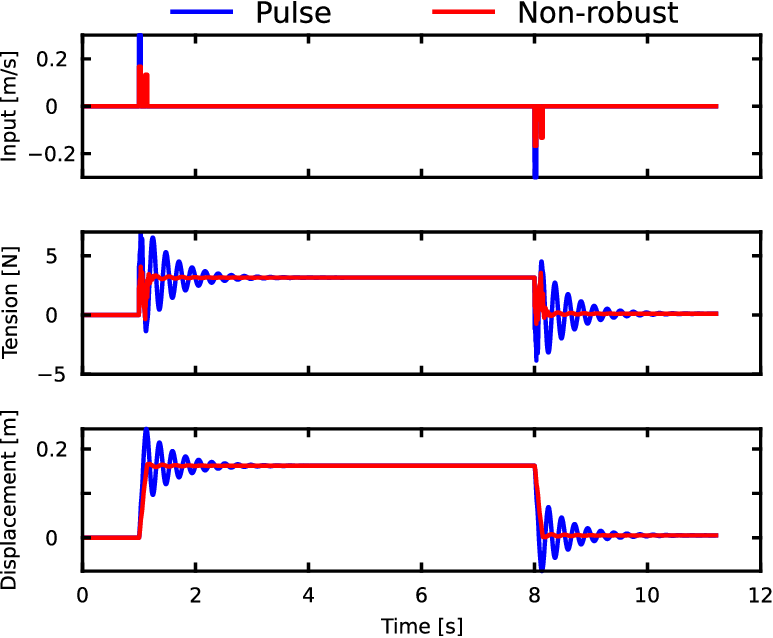}
    \caption{System model response from Eqs.~\eqref{eq:truncated_modal_model_v1}-\eqref{eq:truncated_modal_model_v2} for the velocity driven continuum robot without and with input shaping.}
    \label{fig:12_model_wo_IS_and_with_IS}
\end{figure}

It is well known that for the system model $G(s)$, the complex poles lie at \mbox{$s = -\zeta\omega_\mathrm{n} \pm j\omega_\mathrm{n}\sqrt{1-\zeta^2}$} are those associated with vibration in the transient response. Input shaping modifies the system’s command signal by convolving the input with timed impulses that interact with the system’s natural oscillations. These impulses are timed so that the vibrations excited by each of the delayed and scaled inputs cancel each other out, minimizing residual oscillations of the system response. Essentially, input shaping aligns the input command with the system dynamics to achieve a smooth and vibration-free response. The schematic overview can be seen in Fig.~\ref{fig:4_Block_Diagram_Input_Shaping}.
A single-delay input shaper is expressed as~\cite{singh_optimal_2009}:
\begin{subequations}
\begin{align}
    G_{\mathrm{nonrob}}(s) = A_0 + A_1 e^{-s T},\\
    A_0 = \frac{\exp\left(\frac{\zeta \pi}{\sqrt{1-\zeta^2}}\right)}{1 + \exp\left(\frac{\zeta \pi}{\sqrt{1-\zeta^2}}\right)},\\
    A_1 = 1 - A_0; \quad    T = \frac{\pi}{\omega_d},
\end{align}
\end{subequations}
where $A_0$ and $A_1$ define the amplitudes and $T$ denotes the time delay, also referred as switch time. $\omega_d$ is the damped natural frequency defined as $\omega_d=\omega_n\sqrt{1-\zeta^2}$. This design is referred to as a non-robust TDF. To improve robustness against deviations in system dynamics, a second pair of zeros is placed on the system poles which characterizes the design of a robust TDF as:
\begin{align}
    G_{\mathrm{robust}}(s) = \left(A_0 + A_1 e^{-s T}\right)^2.
\end{align}
This effectively places two identical zeros at the nominal locations of the system’s underdamped poles. This two-impulse design is particularly useful when the identification of the system poles is difficult to obtain. With adding switches in the feedforward structure as described above the maneuver time gets extended. Fig.~\ref{fig:12_model_wo_IS_and_with_IS} illustrates the effect of an input shaper on the system model with a pulse input as described previously.
%
%
%
\section{RESULTS AND DISCUSSION}
\label{sec:results}
\begin{figure}
    \centering
    \includegraphics[width=0.85\linewidth]{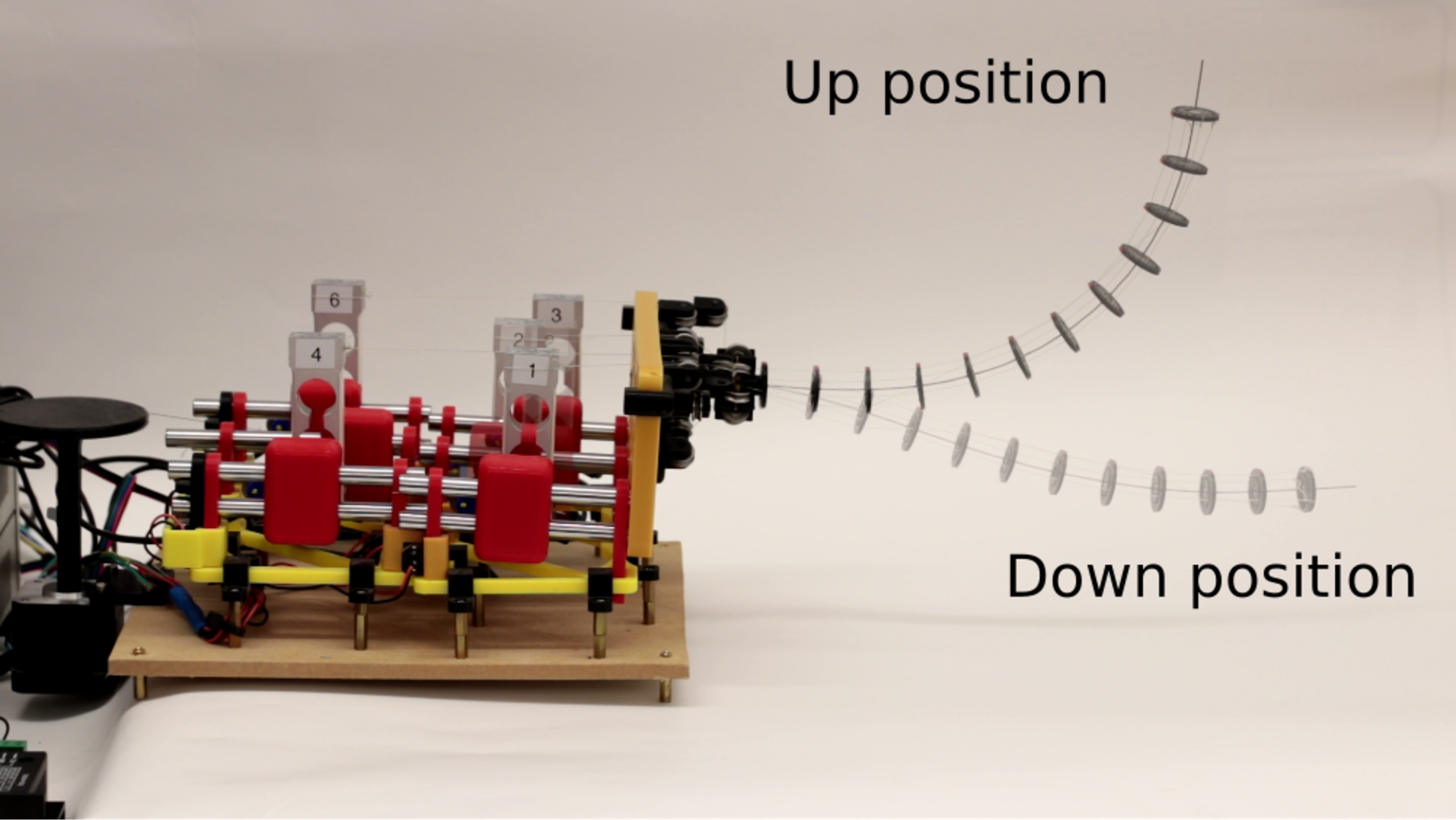}
    \caption{Continuum robot motion shows upward bending when the stepper motor pulls the cable and the return to the downward position when the cable is released.}
    \label{fig:7_continuum_robot_movement}
\end{figure}

\begin{figure}
    \centering
    \includegraphics[width=3.4in]{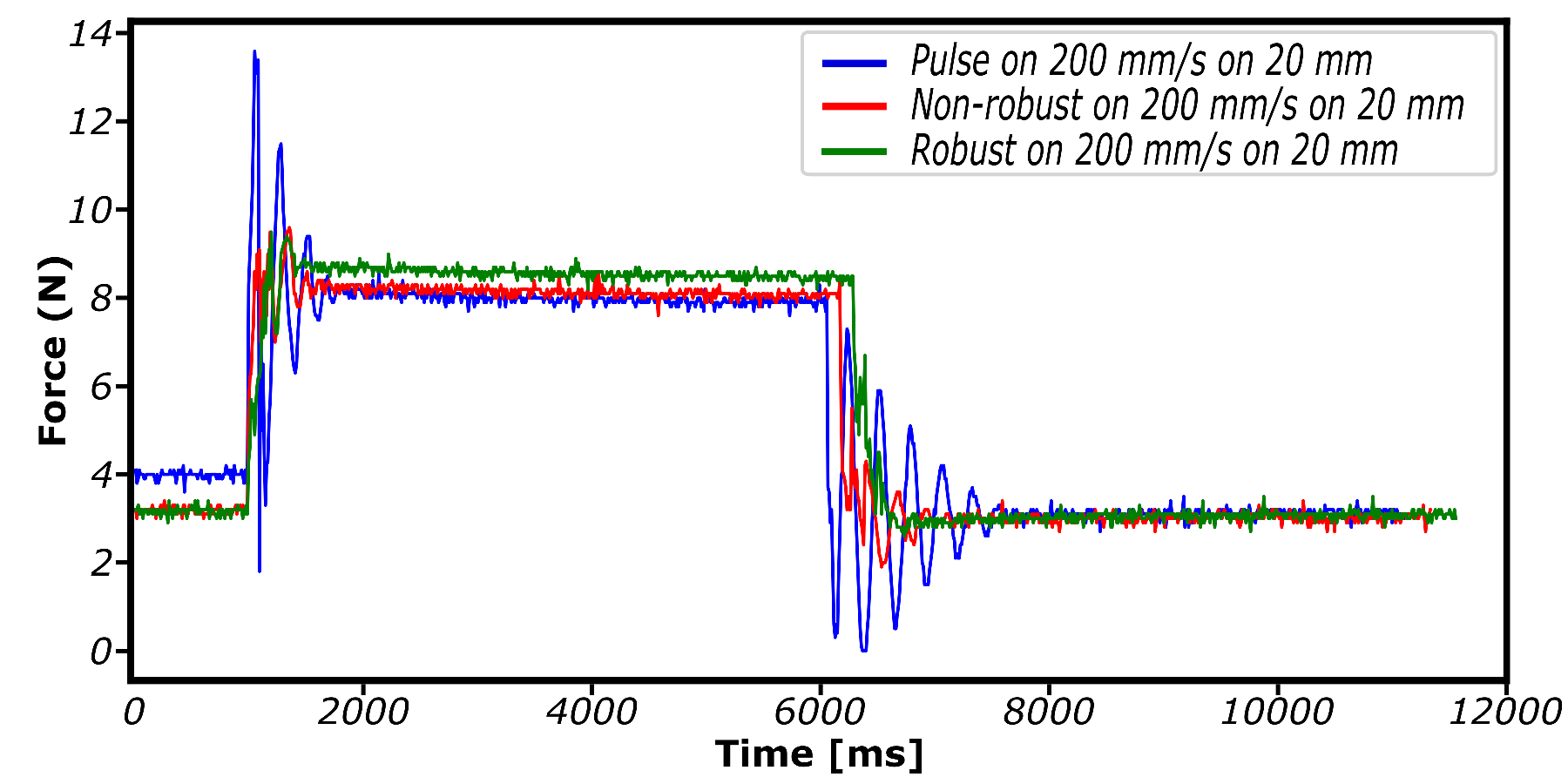}
    \caption{Scenario with $20$ mm displacement at $200$ mm/s velocity.}
    \label{fig:9_20mm_200mms}
\end{figure}

\begin{figure}
    \centering
    \includegraphics[width=3.4in]{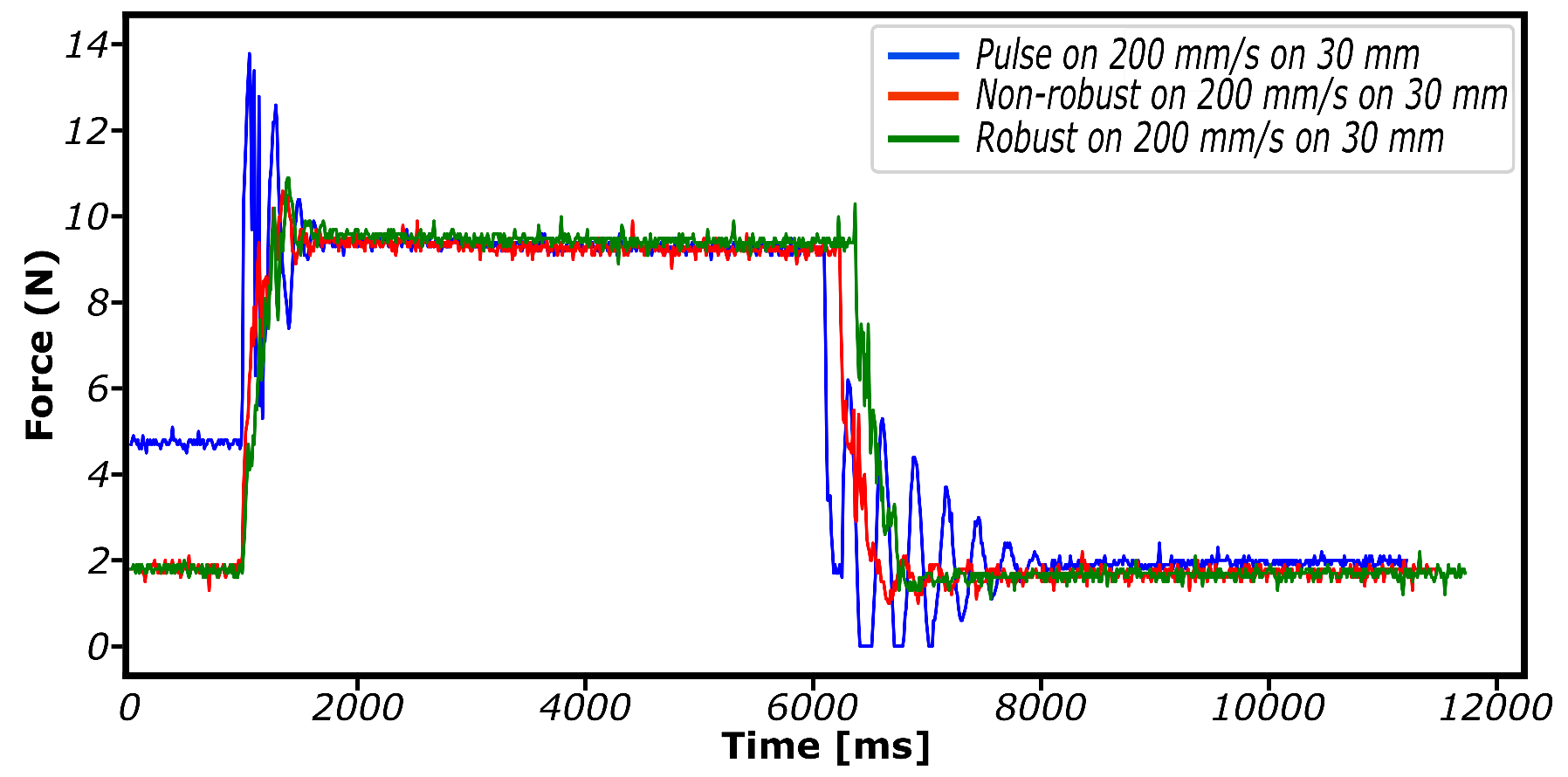}
    \caption{Scenario with $30$ mm displacement at $200$ mm/s velocity.}
    \label{fig:10_30mm_200mms}
\end{figure}

\begin{figure}[t]
    \centering
    \includegraphics[width=3.4in]{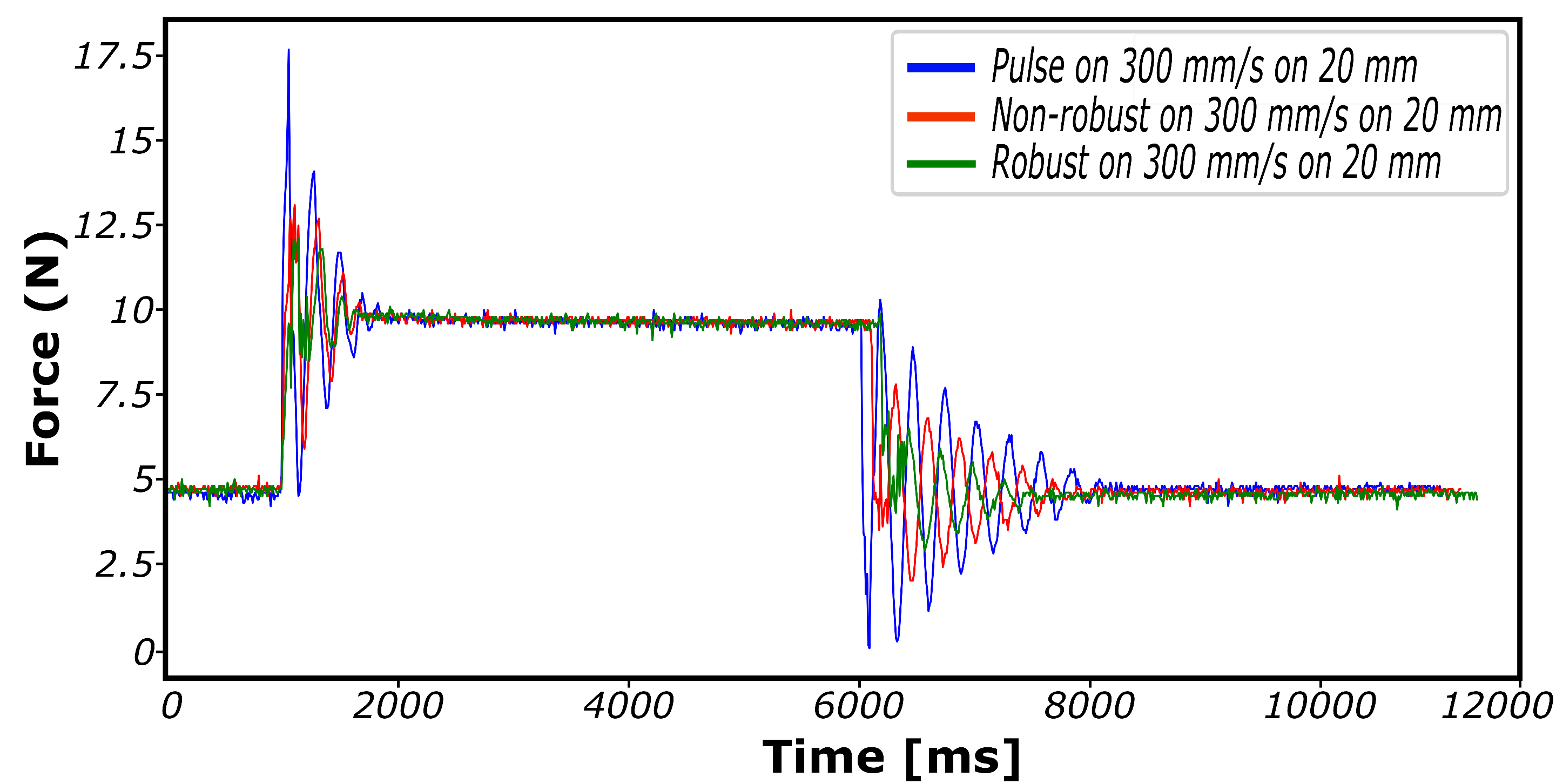}
    \caption{Scenario with $20$ mm displacement at $300$ mm/s velocity.}
    \label{fig:11_20mm_300mms}
\end{figure}

We compare three cases (pulse without input shaping, non-robust TDF, and robust TDF) experimentally on the hardware. 
Each test commands an up–down motion starting at the down position. The same pulsed velocity profile that was used for the system identification is used as the system input, for varying values of the total cable displacement and velocity. The cable tension is recorded over time. The commanded motions are large, as illustrated in Fig.~\ref{fig:7_continuum_robot_movement}, to assess whether the input shaper designs based on the linear system model are effective for the real system. As seen both in the quantitative metrics shown in Table~\ref{tab:metrics} and the tension time responses in Figs.~\ref{fig:9_20mm_200mms} through~\ref{fig:11_20mm_300mms}, the robust TDF generally results in substantial improvements in performance by reducing overshoot and settling time, while the non-robust TDF results in some improvement over the uncompensated response but lesser in effect.

The linearized model does not accurately capture the large deflections of the system. For an imposed 30~mm cable displacement, the model predicted tip displacement of the robot is approximately 0.5~m, which indicates that the system's state is well outside of the range of validity for the linearization. Despite the model inaccuracy in the robot tip displacement predictions, the input shaping remains effective even for large, geometrically nonlinear deformations. The reason for this is that period of the real vibrations at deflected configurations of the robot remains consistent enough with the predicted period of the small vibrations of the linearized model for the input shapers to be effective. 
\begin{table}
\centering
\caption{Quantitative results for pulse, non-robust, and robust input shaping across different displacements and velocities.}
\label{tab:metrics}
\scriptsize 
\renewcommand{\arraystretch}{1.2}
\begin{tabular}{|c|c|c|c|c|c|}
\hline
\textbf{Condition} & \textbf{System} & \textbf{Motion} & \makecell{\textbf{Peak}\\\textbf{(N)}} & \makecell{\textbf{Overshoot}\\\textbf{(\%)}} & \makecell{\textbf{Settling}\\\textbf{Time (ms)}} \\
\hline

\multirow{6}{*}{\makecell{$10$ mm\\$200$ mm/s}}
 & \multirow{2}{*}{Pulse}      & Up   &  $9$ & $153.1$\% & $1223$ \\
 &                              & Down &  $0$ &  $75.9$\% & $1608$ \\ \cline{2-6}
 & \multirow{2}{*}{Non-robust} & Up   &  $7.5$ &  $94.2$\% & $1118$ \\
 &                              & Down &  $0.4$ &  $60.1$\% & $1378$ \\ \cline{2-6}
 & \multirow{2}{*}{Robust}     & Up   &  $6.3$ &  $51.1$\% &  $1002$ \\
 &                              & Down &  $0.8$ &  $44.3$\% & $1348$ \\ \hline

\multirow{6}{*}{\makecell{$20$ mm\\$200$ mm/s}}
 & \multirow{2}{*}{Pulse}      & Up   & $13.6$ & $131.7$\% & $1253$ \\
 &                              & Down &  $0$ & $63.2$\% & $1899$ \\ \cline{2-6}
 & \multirow{2}{*}{Non-robust} & Up   &  $9.6$ &  $27.0$\% & $1014$ \\
 &                              & Down &  $1.9$ & $20.0$\% & $1149$ \\ \cline{2-6}
 & \multirow{2}{*}{Robust}     & Up   &  $9.5$ &  $12.5$\% &  $898$ \\
 &                              & Down &  $2.6$ & $6.1$\% &  $878$ \\ \hline

\multirow{6}{*}{\makecell{$30$ mm\\$200$ mm/s}}
 & \multirow{2}{*}{Pulse}      & Up   & $13.8$ &  $92.9$\% & $1138$ \\
 &                              & Down &  $0$ & $25.9$\% & $2140$ \\ \cline{2-6}
 & \multirow{2}{*}{Non-robust} & Up   & $10.6$ &  $15.3$\% & $1044$ \\
 &                              & Down &  $0.1$ & $7.7$\% & $1211$ \\ \cline{2-6}
 & \multirow{2}{*}{Robust}     & Up   & $10.9$ &  $16.5$\% & $1044$ \\
 &                              & Down &  $1.1$ &  $5.9$\% &  $970$ \\ \hline

\multirow{6}{*}{\makecell{$20$ mm\\$300$ mm/s}}
 & \multirow{2}{*}{Pulse}      & Up   & $17.7$ & $151.6$\% & $1328$ \\
 &                              & Down & $0$ & $96.4$\% & $2475$ \\ \cline{2-6}
 & \multirow{2}{*}{Non-robust} & Up   & $13.1$ &  $65.6$\% & $1182$ \\
 &                              & Down &  $2$ &  $51.3$\% & $2068$ \\ \cline{2-6}
 & \multirow{2}{*}{Robust}     & Up   & $12.1$ &  $43.8$\% & $1139$ \\
 &                              & Down &  $2.9$ &  $31.5$\% & $1567$ \\ \hline

\end{tabular}
\end{table}
\section{CONCLUSIONS}
\label{sec:conclusions}
This paper presented an experimental investigation and comprehensive analysis of input shaping for a cable-driven continuum robot. It was shown that a velocity-driven cable system subjected to a pulse input produces excessive vibrations. To address this, both non-robust and robust input shapers were evaluated under variations in final displacement and cable pulling velocity. While the non-robust shaper already reduced overshoot and shortened settling time compared to the pulse input, the robust shaper consistently delivered superior performance across all tested scenarios, with only a modest increase in maneuver time. These results are significant, as reliable motion control is critical for ensuring good tracking performance in future applications. Building on these findings, future research will extend the framework to more complex motion tasks beyond single-degree-of-freedom point-to-point trajectories.
%
%
%
\bibliographystyle{IEEEtran}
\bibliography{references}
\end{document}